\documentclass[10pt,conference,compsocconf,cmex10]{IEEEtran}
\usepackage[utf8]{inputenc}
\usepackage{microtype}
\usepackage[latexcolors]{xcolor}
\usepackage[hidelinks]{hyperref}
\usepackage{cite}
\usepackage{float}
\usepackage[pdftex]{graphicx}
\usepackage{amsfonts}
\usepackage{amsmath}
\usepackage{amssymb}
\usepackage{mathtools}
\usepackage{fixltx2e}
\usepackage{booktabs}
\usepackage{tikz}
\usepackage{tikzpagenodes}      
\usetikzlibrary{
  arrows,
  arrows.meta,
  automata,
  calc,
  chains,
  math,
  matrix,
  positioning,
  shapes,
  shapes.geometric,
}
\pgfdeclarelayer{background}
\pgfdeclarelayer{foreground}
\pgfsetlayers{background,main,foreground}
\tikzset{
  x=1em,
  y=1em,
  >=stealth',
  node distance=2em,
}

\begin{document}
\title{Bridging the Gap between Semantics and Multimedia Processing}
\author{%
  \IEEEauthorblockN{%
    Marcio Ferreira Moreno\IEEEauthorrefmark{1},
    Guilherme Lima\IEEEauthorrefmark{1},
    Rodrigo Santos\IEEEauthorrefmark{1},
    Roberto Azevedo\IEEEauthorrefmark{2},
    and Markus Endler\IEEEauthorrefmark{3}}
  \IEEEauthorblockA{%
    \IEEEauthorrefmark{1}IBM Research, Brazil\\
    mmoreno@br.ibm.com, guilherme.lima@ibm.com, rodrigo.costa@ibm.com}
  \IEEEauthorblockA{%
    \IEEEauthorrefmark{2}EPFL, Switzerland\\
    roberto.azevedo@epfl.ch}
  \IEEEauthorblockA{%
    \IEEEauthorrefmark{3}PUC-Rio, Brazil\\
    endler@inf.puc-rio.br}
}
\maketitle

\begin{abstract}
  In this paper, we give an overview of the semantic gap problem in
  multimedia and discuss how machine learning and symbolic AI can be
  combined to narrow this gap.  We describe the gap in terms of a classical
  architecture for multimedia processing and discuss a structured approach
  to bridge it.  This approach combines machine learning (for mapping
  signals to objects) and symbolic AI (for linking objects to meanings).
  Our main goal is to raise awareness and discuss the challenges involved in
  this structured approach to multimedia understanding, especially in the
  view of the latest developments in machine learning and symbolic AI.
\end{abstract}


\section{Introduction}
\label{sec:1:intro}

A classic problem in multimedia representation and understanding is the
\emph{semantic gap} problem~\cite{Sikos-L-F-2017}.  It states that there is
a big representational gap between the audiovisual signals that compose
multimedia objects and the concepts represented by these signals.  For
instance, the dominant color and movement trajectory of a given set of
pixels in a video clip, which are low-level characteristics of the clip,
usually do not provide much information about the \emph{meaning} of the set
of pixels---at least not to computers.  But recent developments in
artificial intelligence (AI) are changing that.

Backed by large training datasets, current machine learning methods are able
to extrapolate complex patterns from low-level multimedia data.  These
patterns are embodied in trained models which can be used to classify or
identify persons and objects with reasonable speed and accuracy in images,
audio clips, and to a lesser extent video clips~\cite{Zhang-W-2019}.

But being able to identify persons and objects in multimedia data only
solves half of the problem.  To emulate human cognition and truly understand
a scene---for instance, to determine who is doing what and the consequences
of those actions---computers need additional information: they need common
sense knowledge and domain knowledge, and also the capacity to infer new
knowledge from preexisting knowledge.  This is where symbolic AI comes in.
The basic idea of symbolic AI is to describe the world, its entities, and
their relationships using a formal language and to develop efficient
algorithms to query and deduce things from these formal descriptions.

In this paper, we give an overview of the semantic gap problem in multimedia
and discuss how machine learning (ML) and symbolic AI can be combined to
narrow this gap.  More specifically, we highlight the fact that what we call
\emph{the} semantic gap consists of many gaps which exist between the
various layers of multimedia representation.  A structured approach to
tackle the gap as a whole is thus to tackle each of these smaller gaps
individually, through a combination of ML (for mapping signals to objects)
and symbolic AI (for linking objects to meaning).

Our main goal here is to raise awareness and discuss the challenges involved
in this structured approach to multimedia understanding, especially in view
of the latest developments in ML and symbolic AI.

The rest of the paper is organized as follows.  In
Section~\ref{sec:2:why-symb-ai}, we define what we mean by ``symbolic AI''
and justify why we need it.  In Section~\ref{sec:3:many-gaps}, we describe
how the semantic gap problem is distributed among the various layers of
multimedia representation, and discuss a structured approach for multimedia
understanding.  In Section~\ref{sec:4:challenges}, we discuss the challenges
involved in such a structured approach.  Finally, in
section~\ref{sec:5:final}, we draw our conclusions.



\section{Why We Need Symbolic AI}
\label{sec:2:why-symb-ai}

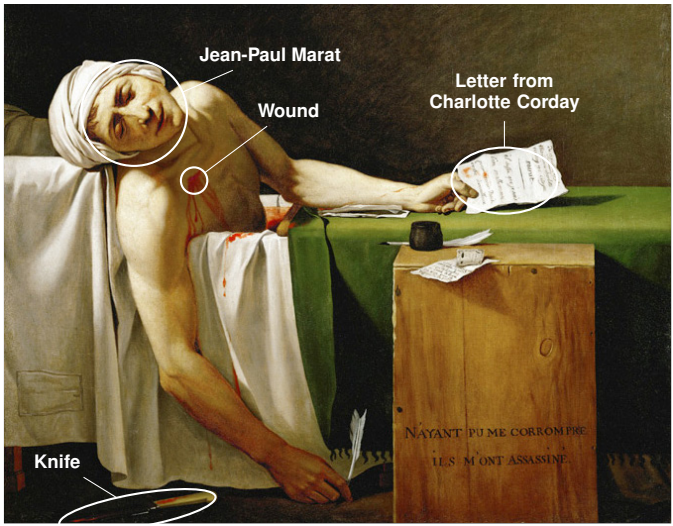
\begin{figure}[t]
  \centering
  \begin{tikzpicture}[node distance=1em,
    w/.style={draw,white,thick,inner sep=0,outer sep=0},
    l/.style={white,font=\sffamily\bfseries\scriptsize}]
    \node[draw,anchor=south west,inner sep=0pt](marat) at (0,0)
    {\includegraphics[width=\columnwidth]{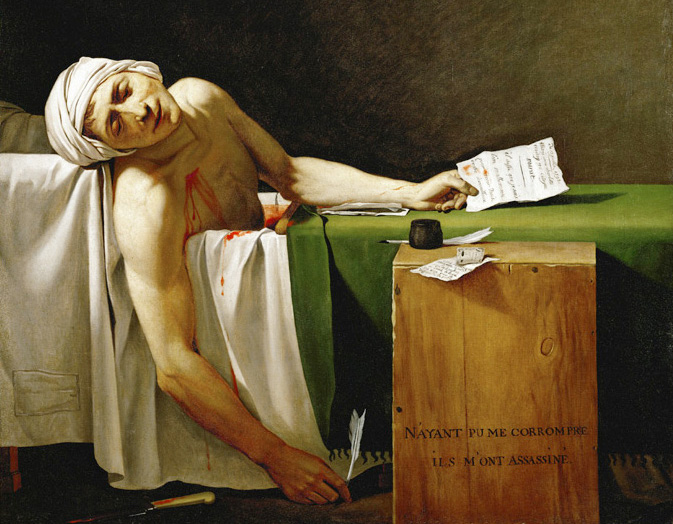}};
    \begin{scope}[x={(marat.south east)},y={(marat.north west)}]
      \clip(0,0) rectangle (1,1);
      \node[w,shape=ellipse,minimum width=6em,
      minimum height=1.25em,rotate=10](knife)at (.2,.025){};
      \node[l](knifelbl) at (.08,.12){Knife};
      \draw[white](knifelbl) to (knife);
      \node[w,shape=ellipse,minimum width=4em,minimum height=4em](head)
      at (.195,.79){};
      \node[l,above right=of head,yshift=-.5em](headlbl){Jean-Paul Marat};
      \draw[white](headlbl) to (head);
      \node[w,shape=ellipse,minimum width=1em,minimum height=1em](wound)
      at (.285,.66){};
      \node[l,above right=of wound,xshift=1em,yshift=1em]
      (woundlbl){Wound};
      \draw[white](woundlbl) to (wound);
      \node[w,shape=ellipse,minimum width=4em,minimum height=2.5em](letter)
      at (.75,.66){};
      \node[l,text width=8em,align=center,above=of letter](letterlbl)
      {Letter from\\Charlotte Corday};
      \draw[white](letterlbl) to (letter);
    \end{scope}
  \end{tikzpicture}
  \abovecaptionskip=0pt
  \belowcaptionskip=0pt
  \caption{\emph{The Death of Marat} (detail), by Jacques-Louis
    David, 1793.  (WikiMedia)}
  \label{fig:marat}
\end{figure}

To illustrate the kind of applications enabled by the combination of
symbolic AI with machine learning and multimedia consider
Figure~\ref{fig:marat}.

Suppose we are given this picture and suppose the only thing we know about
it is what we can infer from the image.  We can see it depicts Jean-Paul
Marat (assuming we can identify him), a stab wound, a blood-tainted knife,
and a letter addressed to him and signed by Charlotte Corday (assuming we
can read the contents of the letter).  The analogy here is that we have
extracted these information---or \emph{facts}---using pattern matching.
Although such basic facts allow us to perform simple computational tasks,
such as keyword-based image classification and search, they are not enough
to understand the image.

To truly \emph{understand} what is being depicted in Figure~\ref{fig:marat}
we need more than basic facts.  We need (1)~general knowledge about the
world, (2)~specific knowledge about the persons named, and (3)~the capacity
to combine general and specific knowledge with the facts extracted from the
image in order to infer new facts.

Now suppose we are given~(1), (2), and (3).  From our general knowledge of
the world, and possibly by further analyzing the image, we can assert with
high confidence that Marat is holding the letter and that he has a stab
wound on the chest.  From this and from the blood-tainted knife depicted
below him, we might infer that the depicted knife is the object that caused
the wound.  Since knifes are not autonomous beings, we might also conclude
that someone (possibly himself) stabbed Marat in the chest.  But who and
why?

To answer these questions we will need more information.  Suppose we are
told that Marat was a journalist and political agitator, and one of the
leaders of a radical political faction in the Reign of Terror period of the
French Revolution (c.~1793).  Suppose we are also told that Charlotte
Corday, who signed the letter, was a declared political enemy of Marat---she
blamed him for a number of killings in Paris and other cities and believed
that he was a grave threat to the French Republic.  Under the light of these
new facts, we can conclude that Figure~\ref{fig:marat} looks like the scene
of a political murder.

By combining this conclusion with the additional fact that Charlotte Corday
is known to have murdered Jean-Paul Marat with a knife while he was in his
bathtub, holding a letter from her, we can infer with a high degree of
confidence that Figure~\ref{fig:marat} must be a graphical representation of
this incident, that is, of the politically motivated assassination of
Jean-Paul Marat by Charlotte Corday.

The derivation of this last fact from the visual patterns of
Figure~\ref{fig:marat} has only been possible because we have had access not
only to basic facts extracted from the image, but also to facts about the
world (common sense knowledge) and about the depicted objects and persons
(domain knowledge), and because we could combine all of these facts and make
inferences.

One of the main goals of symbolic AI is to enable the representation and
manipulation of pieces of knowledge by computers in ways that resemble or
emulate the kind of manipulations performed by humans---manipulations
similar to the considerations that enabled us to determine the true meaning
of Figure~\ref{fig:marat}.  The combination of this capacity with multimedia
opens up many possibilities.  For instance, the Marat's murder example is an
application of automated image understanding.  Two related applications are
video understanding and audio understanding, which are often more complex as
they involve the extraction of temporal information.

Other applications of symbolic AI to multimedia include the semantic
retrieval, classification, recommendation, and inspection of multimedia
data---for example, to automatically identify suspicious activity in
surveillance videos, generate age ratings for music and movies, and identify
risk factors for diseases in medical images and videos.





\section{A Structured Approach to Multimedia Understanding}
\label{sec:3:many-gaps}

\subsection{The many semantic gaps}

\begin{figure}[h!]
  \centering
  \begingroup
  \newcommand*\X[1]{\footnotesize\textsl{#1}}
  \begin{tikzpicture}
    \matrix(M)[inner sep=0pt,outer sep=0pt,row sep=-\pgflinewidth,column sep=1em,
    n/.style={draw,text width=10em,minimum height=3em,align=center}]{
      \node[overlay]{(4)};&\node[n](4){Semantics\\\X{object relationships}};\\
      \node[overlay]{(3)};&\node[n](3){Object Labels\\\X{names of objects}};\\
      \node[overlay]{(2)};&\node[n](2){Objects\\\X{combinations of descriptors}};\\
      \node[overlay]{(1)};&\node[n](1){Descriptors\\\X{feature vectors}};\\
      \node[overlay]{(0)};&\node[n](0){Raw Content\\\X{audiovisual signals}};\\
    };
    \begin{scope}[overlay,shorten >=3pt]
      \foreach\i\j in {0/1,1/2,2/3,3/4}{
        \draw[->]($(\i.east)+(1em,0)$)
        to node[right,yshift=-3pt]{$m_\j$} ($(\j.east)+(1em,0)$);
      };
    \end{scope}
  \end{tikzpicture}
  \endgroup
  \caption{Hierarchy of levels between raw content and its full semantics.}
	\label{fig:gap}
\end{figure}
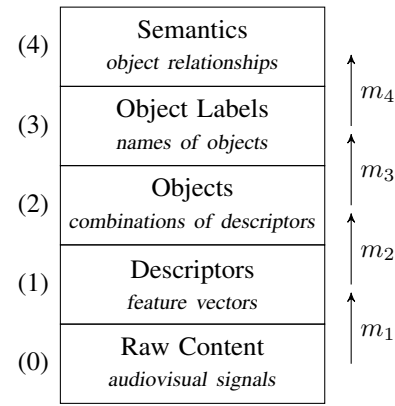

There is a hierarchy of layers of processing and representation separating
raw multimedia data (arrays of bytes) from their semantics (meaning).  The
so-called semantic gap can occur between any two of these layers.  To see
why this is the case, consider Figure~\ref{fig:gap} which depicts the
classic structure of a bottom-up pipeline for extracting semantics from
multimedia content~\cite{Hare2006MindTG}.

In the classical version of the pipeline, the first step, mapping $m_1$ in
Figure~\ref{fig:gap}, consists in extracting feature vectors describing
low-level characteristics of the data, such as color histograms, pixel
intensities, length of sounds, noise ratios, frame rates, etc.  These
feature vectors are then used by mapping $m_2$ as a basis for identifying
objects.

Note that the gap is already present at mapping $m_2$.  The question of how
to extract or segment objects from feature vectors is far from simple,
especially when only classical signal processing methods are considered.  In
the case of Figure~\ref{fig:marat}, the mapping $m_2$ corresponds to the
problem of grouping sets of features in order to identify contours and
objects, for instance, the knife, the person, the letter, etc.
(See~\cite{Brezeale-D-2008,Fu-Z-2011,Khan2014} for a survey of approaches
for tackling this problem for image, audio, and video data.)

Continuing upwards in the pipeline of Figure~\ref{fig:gap}, once different
objects are identified (layer 2), the next problem, mapping $m_3$, is to
assign labels to these objects.  In terms of Figure~\ref{fig:marat}, this
means to assign the labels ``knife'', ``letter'', ``Marat'', and ``wound''
to the appropriate segments of the image.

As before, mapping $m_3$ also has to deal with a gap.  In this case, the gap
between the objects identified by mapping $m_2$ and the labels which should
represent them.  The problem here is that an object might be mapped to
multiple labels, the same label might refer to different objects, and the
choice of label might depend on the context in which the object occurs (for
instance, on its surroundings).

Most of the research related to the problem of the semantic gap in
multimedia is concerned with mappings $m_1$, $m_2$ and $m_3$.  That is, with
bridging the distance between the raw content and the object labels.  Even
though this is crucial for multimedia understanding, as discussed in
Section~\ref{sec:2:why-symb-ai}, it comprises only half of the problem.
After $m_3$, we are sill left with a gap between the labels and their
meaning---a gap that must be tackled by mapping $m_4$.  The idea is that
after $m_4$, in layer 4, we are no longer dealing with objects and labels,
but with meaningful concepts.  In case of Figure~\ref{fig:marat}, a knife
(the sharp object and its properties), Marat (the person and its history),
etc.

We now shift the discussion to AI; to how the latest developments in machine
learning and symbolic representation are contributing to narrow the gaps
between the layers of Figure~\ref{fig:gap}, and what are the challenges in
these areas.

\subsection{Machine learning: From data to objects and labels}

In the last years, deep learning methods for visual object recognition and
speech recognition have achieved unprecedented levels of accuracy, often
surpassing by far that of traditional signal processing
techniques~\cite{Druzhkov-P-N-2016,Francesc-A-2016}.  Given enough data and
computational power, it is possible to train a neural network to effectively
``solve'' a specific pattern recognition task.  The problem is that in the
case of multimedia, the task is hardly specific and we are often dealing
with data that is ambiguous or imprecise.

So, although deep learning has significantly improved our ability to map raw
multimedia content to objects and labels (mappings $m_1$, $m_2$, and $m_3$
in Figure~\ref{fig:gap}), there are still many open problems in this area.
We discuss some of these in Section~\ref{sec:4:challenges}.

\subsection{Symbolic AI: From objects and labels to meanings}

The idea of symbolic AI is to describe the world, its entities, and their
relationships using a formal language.  And then to apply formal queries and
reasoning over these descriptions.  One can argue that the Semantic Web
technologies~\cite{Hitzler-P-2010} are the most successful realization of
this idea, at least in terms of adoption.  The use of ontologies (formal
conceptualizations of particular domains) has become widespread, including
within the multimedia community, where there is a growing interest in using
ontologies to describe and derive higher-level facts from multimedia
data~\cite{Nevatia-R-2004,Saathoff-C-2010,Chaudhury-S-2015,Carbonaro-A-2018}

Symbolic representation and reasoning is the key to bridge the gap between
object labels and their meaning (mapping $m_4$ in Figure~\ref{fig:gap}).
Using an ontology language such as OWL~\cite{W3C-OWL2Overview} (a standard,
Semantic Web technology), we can describe what the labels ``knife'' and
``Marat'' mean---what these entities are, how they behave, and what is their
relation to the rest of the world.  And from these descriptions, through
formal reasoning, we can eventually derive the conclusion that the scene
depicted in Figure~\ref{fig:marat} is that of a political murder.

That said, the semantic description of multimedia data is far from being a
simple problem.  The imprecise and ambiguous nature of multimedia requires
descriptions that allows us to represent fuzziness and probability.
Moreover, these same descriptions should allow us to speak of time and
space, two fundamental dimensions of audiovisual data, and to link
descriptions to the underlying fragments of data they describe.  These and
other limitations of current representation models have backed the
development of hybrid models, such as Hyperknowledge~\cite{hk,Moreno-M-2016}, which
combines multimedia data and its semantic description under a single
framework, and has support for spatial and temporal queries.  We discuss
these and other challenges next.



\section{Challenges}
\label{sec:4:challenges}

There are still many challenges related to semantic gap in multimedia.  We
discuss some of these below.

\subsubsection{Mulsemedia}
Multiple sensorial media (Mulsemedia) is a recent trend that focuses on
enriching traditional multimedia content with modalities that go beyond
audio and video, that is, modalities such as haptic, olfactory,
thermoceptic, etc.  While there are many well-known algorithms for
processing audiovisual signals, the algorithms and tools for processing
mulsemedia are still in their infancy.  Future research should tackle
problems such as extracting meaning from sensorial media and aligning this
meaning with the semantics extracted from traditional media
contents~\cite{Saleme:2018,Naravane2017}.

\subsubsection{Moving objects}
The detection and tracking of moving objects in videos is a well-known
problem in video segmentation~\cite{Yilmaz-A-2006,ciaparrone2019deep}.
Related to this problem and still a largely unexplored area of research, is
the problem of describing such moving objects semantically in way that
captures the notion that they have an identity even though they can possibly
change through time.

For instance, when watching a video clip a person can easily determine that
a woman walking on a sidewalk which gets temporarily occluded by a passing
car is still the same woman after the car passes.  Similarly, a child that
puts on a mask is still the child, but now a masked one.  The notion of time
and change complicates not only the problem of identifying objects in
audiovisual signals but also the problem of describing these objects and
their changes---descriptions which might require a non-monotonic model of
truth (things that are true now might not be true in the future).


\subsubsection{Real-time reasoning}
Real-time reasoning over audiovisual streams is another challenging topic.
In general, approaches based on deep learning introduce non-negligible
delays that make them unsuitable for applications that require near
real-time performance.  Things are even worse on the symbolic side, where
the complexity of reasoning tasks is often related to the size of the
knowledge base.  Real-time reasoning over multimedia data requires
processing and correlating spatial (intra-frames) and temporal
(inter-frames) data with corresponding symbolic descriptions on the fly.
This is pretty much an open problem which has to be investigated by the
community.





\section{Final Remarks}
\label{sec:5:final}

In this paper, we discussed the problem of the semantic gap in multimedia
representation in the view of the latest developments in AI.  We described
this gap in terms of a classical architecture for multimedia processing and
discussed how the gap is distributed among the layers of this architecture.
We noted that a promising approach for bridging this gap is to do so in a
structured manner through a combination of machine learning (for mapping
signals to objects) and symbolic AI (for linking objects to meanings).  We
discussed the current challenges involved in this approach and also future
challenges, which will require further advances in IA methods and
technologies.



\IEEEtriggeratref{5}
\bibliographystyle{IEEEtran}
\bibliography{paper}
\end{document}